%% file: root.tex
\documentclass[letterpaper, 10 pt, conference]{ieeeconf}
\IEEEoverridecommandlockouts                             
\usepackage{tabularx} 
\usepackage{adjustbox} 
\usepackage{subfigure}
\usepackage{booktabs}
\usepackage{csquotes}
\usepackage{graphicx}
\usepackage{verbatim}
\usepackage{fancybox}
\usepackage{amsmath}
\usepackage{multirow}
\usepackage{tabularx}
\usepackage{cleveref}
\usepackage{tcolorbox}
\let\labelindent\relax

\usepackage{enumitem}
\usepackage{balance}
\usepackage{flushend}
 \usepackage{pgfplots}
\usepackage{algorithm}
\usepackage{algpseudocode}

\title{\LARGE \bf
Systematic Evaluation of Initial States and Exploration-Exploitation Strategies in PID Auto-Tuning: A Framework-Driven Approach Applied on Mobile Robots}

\author{Zaid Ghazal$^{1}$, Ali Al-Bustami$^{2}$, Khouloud Gaaloul$^{1}$ and Jaerock Kwon$^{2}$
\thanks{$^{1}$Z. Ghazal and K. Gaaloul with the Department of Computer Information Systems, Collage of Engineering and Computer Science, University of Michigan-Dearborn, 48128 USA
        {\tt\small zghazal@umich.edu}, {\tt\small kgaaloul@umich.edu}}%
\thanks{$^{2}$A. Al-Bustami and J. Kwon with Department of Electrical and Computer Engineering, Collage of Engineering and Computer Science, University of Michigan-Dearborn, 48128 USA
        {\tt\small abustami@umich.edu}, {\tt\small jrkwon@umich.edu}}%
}

\begin{document}

\maketitle
\thispagestyle{empty}
\pagestyle{empty}

\begin{abstract}

PID controllers are widely used in control systems because of their simplicity and effectiveness. Although advanced optimization techniques such as Bayesian Optimization and Differential Evolution have been applied to address the challenges of automatic tuning of PID controllers, the influence of initial system states on convergence and the balance between exploration and exploitation remains underexplored.  Moreover, experimenting the influence directly on real cyber-physical systems such as mobile robots is crucial for deriving realistic insights. In the present paper, a novel framework is introduced to evaluate the impact of systematically varying these factors on the PID auto-tuning processes that utilize Bayesian Optimization and Differential Evolution. Testing was conducted on two distinct PID-controlled robotic platforms, an omnidirectional robot and a differential drive mobile robot, to assess the effects on convergence rate, settling time, rise time, and overshoot percentage. As a result, the experimental outcomes yield evidence on the effects of the systematic variations, thereby providing an empirical basis for future research studies in the field.
\end{abstract}


\section{Introduction}
\label{sec:introduction}
\input{Sections/Introduction}

\section{Background}
\label{sec:Background}

\input{Sections/Problem}

\section{Methodology}
\label{sec:methodology}

\input{Sections/Methodology}

\section{Evaluation}
\label{sec:evaluation}
\input{Sections/new_eval}

\section{Related Works}
\label{sec:related-works}
\input{Sections/Related}
\section{Conclusion}
\label{sec:conclusion}
\input{Sections/Conclusion}



{
\footnotesize
\bibliographystyle{new_IEEEtran}
\bibliography{./IEEEexample}
}

\end{document}

%% file: Sections/Introduction.tex
Auto-tuning \cite{what-is-auto-tuning} \cite{Automatic-tuning-of-optimum-PID-controllers}  is a method for automatically adjusting parameters in complex systems to achieve optimal performance. This is especially valuable where manual tuning is impractical, such as in robotic applications with large configuration spaces and intricate parameter interdependencies~\cite{auto-tuning}. A key domain for auto-tuning is the Proportional-Integral-Derivative (PID) controller~\cite{what-is-PID}, a widely adopted industrial control method prized for its simplicity and effectiveness~\cite{pid_control_for_linear_and_nonlinear_industrial_processes_2023}. PID controllers adjust the control input based on proportional, integral, and derivative terms of the error signal, with each gain (\(K_p, K_i, K_d\)) acting on current, accumulated, or changing error, respectively~\cite{control_pid}. 

\begin{figure}
    \includegraphics[width=1.0\linewidth]{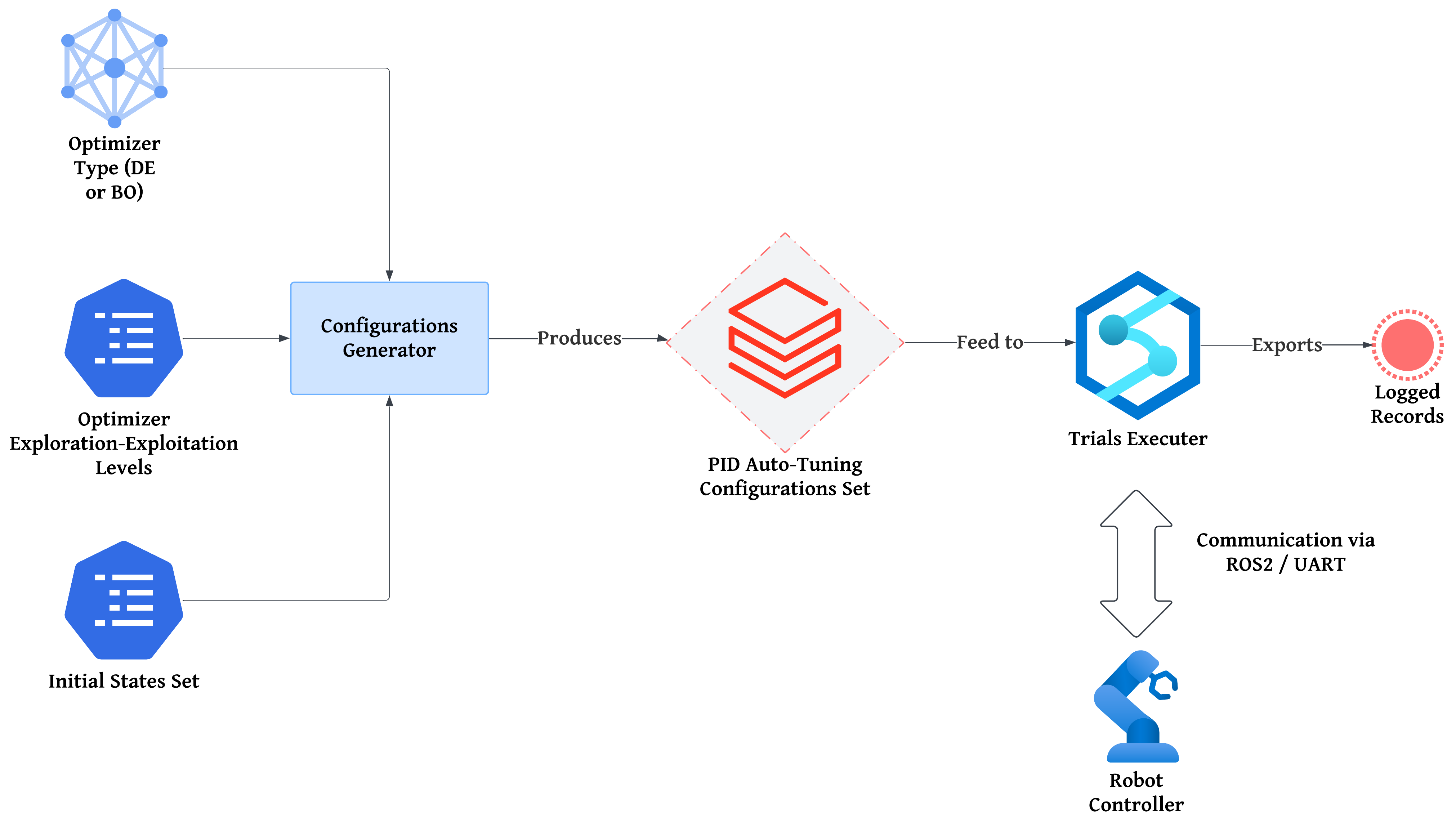}
    \caption{\textbf{Overview of the Proposed PID Auto-Tuning Framework.}
The diagram shows two main modules: a Configurations Generator that produces tuning trials by pairing different initial PID states with exploration–exploitation levels, and a Trials Executer that applies these configurations using Bayesian Optimization and Differential Evolution in mobile robotics experiments.
}
    \label{fig:framework}
\end{figure}

In mobile robotics~\cite{what-is-mobile-robot}, PID controllers serves for critical tasks like navigation and obstacle avoidance \cite{pid_tuning_methods_1} \cite{pid_tuning_methods_3} \cite{pid_tuning_methods_4}. Given the complexity and dynamic nature of robotic systems, numerous optimization-based auto-tuning methods have been proposed to overcome the limitations of classical tuning approaches. Examples include real-time least-squares estimation with a forgetting factor~\cite{Auto-tuning-PID-module-of-robot-motion-system}, metaheuristics such as the Firefly Algorithm (FF)~\cite{what-is-firefly} and Genetic Algorithm (GA)~\cite{what-is-geneticAlgorithm}, and more specialized algorithms like the Bat Algorithm with Mutation (BAM), Social Spider Optimization (SSO), and Particle Swarm Optimization (PSO). These methods often outperform classical approaches in terms of overshoot, steady-state error, and settling time~\cite{Metaheuristics-Algorithm-for-Tuning-of-PID-Controller, bio-inspired-optimization}.
Differential Evolution \cite{differential-evolution-1997} further stands out for its robust population-based search that can effectively handle noisy, time-varying problems and avoid premature convergence~\cite{DE-and-Ga}. Bayesian Optimization, on the other hand, employs probabilistic models (e.g., Gaussian Processes) to guide a more efficient exploration--exploitation balance~\cite{garnett_bayesoptbook_2023}, outperforming alternative methods like BAM and FF in speed and accuracy, especially for black-box functions~\cite{weissenbacher_max_2023}. 

While advanced metaheuristic and surrogate model-based techniques such as Bayesian Optimization and Differential Evolution have been widely applied for PID controller auto-tuning in Cyber-Physical Systems (CPS) such as mobile robots \cite{comparative_study_of_optimal_tuning_pid_controller_for_manipulator_robot_2023} \cite{yongju_pak__2022}  \cite{najah_yousfi_allagui__2021}  \cite{nur_aisyah_syafinaz_suarin__2019} \cite{EvolutiveTuningOptimization}, a clear gap remains in the literature regarding the impact of initial system states and the balance between exploration and exploitation on the convergence and performance of these mobile robotic systems.

This paper introduces a novel framework, shown in Figure \ref{fig:framework}, that streamlines the integration of various exploration-exploitation levels and initial states to generate a unique set of different configurations for Bayesian and Differential Evolution optimizers, the backbone of the auto-tuning process. This enables finding the most suitable configuration-optimizer pair, ensuring that the auto-tuning process yield optimal PID control performance. To demonstrate its practical utility, experiments were conducted on two different robotic platforms, a differential-drive mobile robot and an omnidirectional robot, with each executing PID controlled 90-degree in-place rotations while adhering to predefined overshoot percentage and rise time constraints, and aiming to minimize settling time. Moreover, the experimentation seeks to answer the following Research Questions (RQs): (1)  How does the exploration-exploitation trade-off affect PID auto-tuning convergence across different robot types, in terms of settling time and convergence percentage; (2) How does the initial state impact PID auto-tuning outcomes, specifically settling time and convergence percentage, across various robot types; (3) How do BO and DE differ when used for PID auto-tuning, given the variation in the exploration-exploitation levels and initial states.

\textbf{Structure.} Section~\ref{sec:Background} defines the background. Section~\ref{sec:methodology} presents our framework for the study methodology and experimentation. Section~\ref{sec:evaluation} presents an evaluation of our study. Section~\ref{sec:related-works} compares with the related work. Finally, Section~\ref{sec:conclusion} concludes our paper.



%% file: Sections/Problem.tex
\label{sec:Background}

Recent research has demonstrated that advanced optimization techniques hold great promise for automating controller tuning in robotics. For example, Ribeiro et al. applied BO to tune visual servo and computed torque controllers within a reinforcement learning framework, showcasing BO’s efficiency in handling complex, high-dimensional tuning problems \cite{Ribeiro2021}. Similarly, d’Elia et al. addressed whole-body control challenges by developing an automatic tuning and selection methodology for controllers in robots with multiple degrees of freedom, emphasizing the benefits of systematic parameter adjustment for robust performance \cite{dElia2022}. Van Diggelen et al. further contributed by comparing BO and DE on evolvable morphologies, highlighting the varying performances of these optimization strategies under changing robotic configurations \cite{vanDiggelen2023}. Also relevant to these works, Khosravi et al. demonstrated a performance-driven cascade controller tuning approach using BO, where explicit performance metrics guide the tuning process \cite{Khosravi2020}, whereas Milián et al. proposed a multi-stage tuning framework to reduce the computational cost of BO, thereby enhancing its applicability in real-time, resource-constrained environments \cite{Milian2024}.

A critical aspect shared by these approaches is the balance between exploration and exploitation, which is a trade-off that directly impacts the optimization process’s efficiency and success. Addressing this issue, Candelieri et al. proposed a novel acquisition function that adaptively modulates the uncertainty bonus via a Pareto analysis framework, effectively balancing exploration and exploitation choices within BO \cite{Candelieri2023}. In the evolutionary optimization arena, Sá et al. highlighted a key limitation of standard DE: when the entire population is trapped in a local minima, the algorithm’s ability to search globally is severely compromised \cite{SaEtAl2021}. They introduced modifications to the mutation and crossover operators to explicitly adjust the exploration--exploitation balance. Taking an inverse optimization perspective, Sandholtz et al. developed a probabilistic framework to infer human acquisition functions from observed behavior in sequential optimization tasks, revealing that humans often favor exploration more than standard models predict, which motivates the augmentation of acquisition strategies \cite{Sandholtz2022}. Moreover, Zhang et al. introduced a selective-candidate framework with a similarity selection rule, which generates multiple candidate solutions per individual and selects the final candidate based on both fitness and Euclidean distance \cite{Zhang2020}. This explicit adaptive control of exploration and exploitation has been shown to enhance performance across diverse benchmark problems.

This work presents a unified framework for PID auto-tuning that bridges the gap between theoretical optimization techniques and practical robotic control in mobile robotics. Unlike previous studies that examine Differential Evolution (DE) and Bayesian Optimization (BO) separately or under less systematic conditions, the proposed framework integrates both methods while dynamically adjusting initial PID states and the balance level between exploration and exploitation. This approach generates a comprehensive set of unique configurations that streamline the tuning process and ensure the final PID gains deliver optimal control performance. To validate, experiments were conducted on differential-drive and omnidirectional robots performing 90-degree in-place rotations under predefined overshoot and rise time constraints while minimizing settling time, thereby showing the joint influence of these factors on convergence dynamics and overall performance.

Additionally, in this section we define key terms used in the paper.

\begin{itemize}
    \item {Setpoint}: The desired or target value that a system aims to reach and maintain. In this study, the setpoint value is 90-degrees.

    \item{Exploration-Exploitation Trade-off}: A principle in optimization algorithms that balances exploring new areas of the solution space (exploration) and refining known good areas (exploitation) to find the optimal solution efficiently.

    \item{Initial States}: A set of starting gain values for the PID controller, characterized by different values of $K_p$, $K_i$, and $K_d$. These initial states influence the control system's response, optimization convergence speed, and avoidance of local minima.

    \item{Settling Time}: The time required for the system to stabilize within 5\% of the desired setpoint.

    
    
    
    

    \item{Optimal Gain Values}: The $K_p$, $K_i$, and $K_d$ gains that minimize the settling time (the optimization objective) while keeping the overshoot percentage and rise time within pre-specified ranges, known as constraints.


    \item{Trial}: An experimental run in which an optimizer (DE or BO) is used to tune the PID controller's gain values ($K_p$, $K_i$, $K_d$) for achieving a 90-degree in-place rotation, using specific configurations. The trial ends when the optimization objective is met or when the maximum number of iterations is reached. A detailed description is provided in Section \ref{sec:methodology}

    \item{Iteration}: A single step in a trial where the optimizer adjusts the PID gains and evaluates system performance, refining the solution toward the optimization objective.

\end{itemize}

%% file: Sections/Methodology.tex
Presented in Figure \ref{fig:framework}, the proposed framework orchestrates the PID auto-tuning using two primary components: a Trials Generator, which merges initial states with exploration-exploitation levels to form varied and unique configurations for auto-tuning, and a Trials Executer, which applies these configurations to carry out the process and yield PID gains.

\subsection{First Component: Configurations Generator}
\label{ssec:init_states}

\textbf{Overview.} The Configurations Generator is responsible for producing all unique auto-tuning configurations. It takes two sets as input (1) \emph{Initial States}, each representing a distinct \((K_p, K_i, K_d)\) combination, and (2) \emph{Exploration-Exploitation Levels}, which dictate how aggressively or conservatively the algorithm searches the solution space. By combining them, the generator creates a comprehensive set of unique configurations. In the rest of the paper these configurations sometimes mentioned under the name "trials", as each configuration represents a unique trial to be executed.

\textbf{Initial States.} We define two initial states, chosen to represent diverse tuning scenarios:
\begin{enumerate}
    \item \textbf{Initial State 1 (High P, Low I, Low D):}  A strong proportional response, leading to rapid corrections but increased risk of overshoot and instability.
    \item \textbf{Initial State 2 (High P, Low I, High D):} Balances aggressive proportional action with stronger damping from the derivative term, reducing overshoot and oscillations.
\end{enumerate}
A scenario with low gains across all terms was excluded because it provides insufficient corrective action and is generally ineffective. Additionally, the integral term remains low to mitigate integrator wind-up, which can cause pronounced overshoot and prolonged settling times~\cite{what-is-PID}.

\textbf{Exploration-Exploitation Levels (EELs).} The following EELs guide the search strategy by controlling how much the optimization algorithm explores new regions or exploits known good solutions:
\begin{itemize}
    \item \textbf{Balanced:} Balances exploration and exploitation, helping avoid local optima while steadily progressing toward the global optimum.
    \item \textbf{Exploration-Focused:} Emphasizes discovering new areas of the parameter space, reducing the risk of premature convergence.
    \item \textbf{Exploitation-Focused:} Concentrates on refining known promising solutions more aggressively.
\end{itemize}

\textbf{Producing the auto-tuning configurations using Initial States and EEL Sets .} The Cartesian Product (CP) of these two sets generates all possible trial configurations. If \(A\) represents the set of EELs and \(B\) represents the set of initial states, then the CP \(A \times B\) is defined in Equation~\ref{eq:cp}:

\begin{equation}
\label{eq:cp}
A \times B = \{ (a, b) \mid a \in A \text{ and } b \in B \}.
\end{equation}

In this study, \(A\) has 3 EELs and \(B\) has 2 initial states, resulting in 6 unique ordered pairs for each optimizer. Each pair constitutes a specific trial, defined by one initial PID state and one exploration-exploitation strategy. These trials ensure diverse coverage of the parameter space and optimization behaviors.

\subsection{Second Component: Trials Executer}

\begin{figure*}[t]
    \centering
    \includegraphics[width=0.665\textwidth]{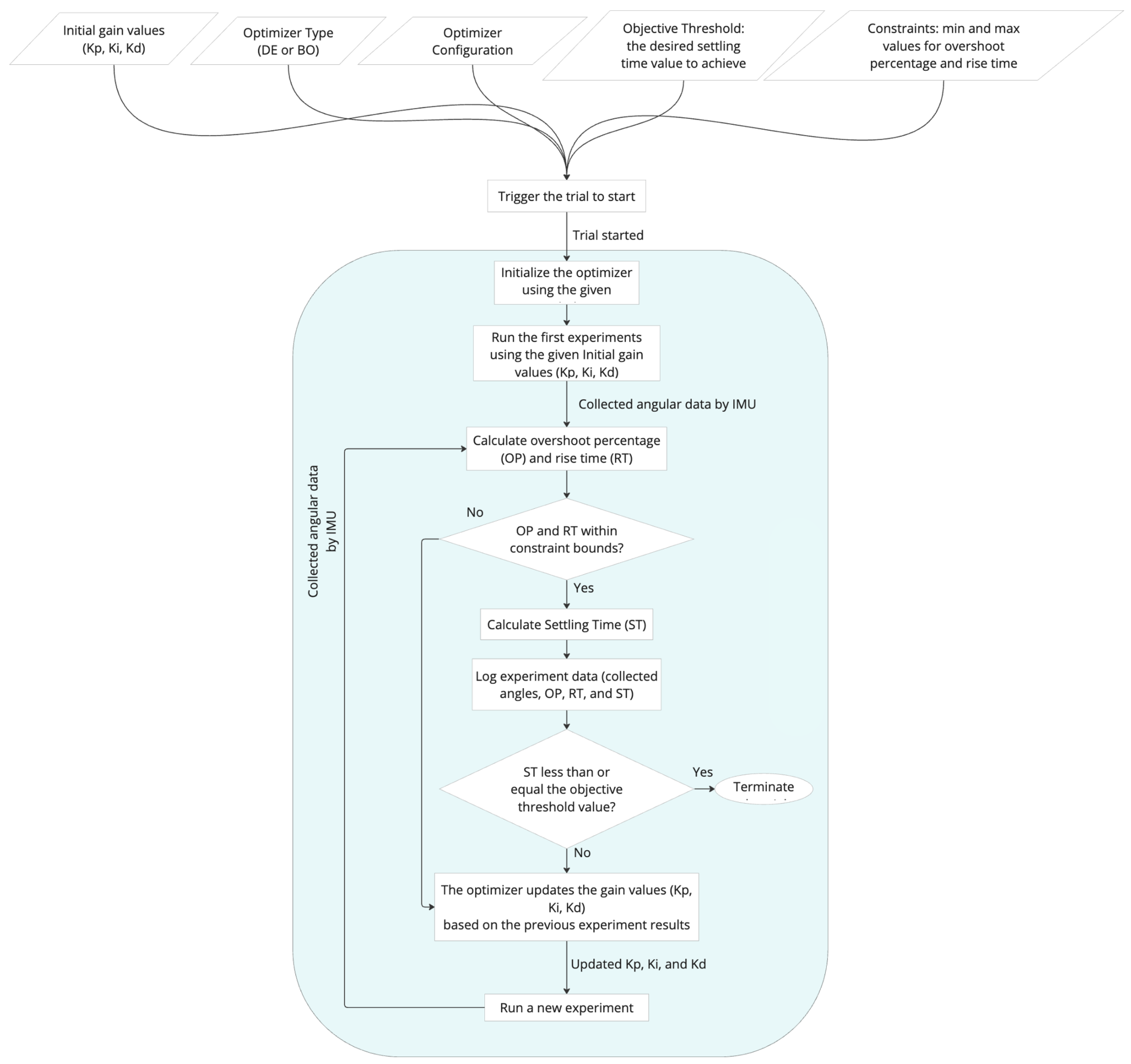}
    \caption{\textbf{Workflow of the Trials Executer.}
 The diagram outlines the process by which each auto-tuning configuration (trial) is sequentially executed using both Bayesian Optimization and Differential Evolution on mobile robotic platforms.
}
    \label{fig:trial-workflow}
\end{figure*}

Trials executer is responsible of conducting each generated trial and collecting results data. Figure \ref{fig:trial-workflow} presents the trials executer workflow.

First, the trials are fed one-by-one to be executed. The executer takes mainly the following:
\begin{itemize}
    \item \textit{Optimizer Type}: type of the optimization algorithm, either DE or BO.
    \item \textit{Objective Threshold}: the target value for the settling time. If this threshold is met, the trial will be terminated.
    \item \textit{Constraints}:  specify minimum and maximum values for both overshoot percentage and rise time. 
    \item \textit{Generated Configuration}: The specific configuration generated by the Configurations Generator, including the initial PID state and exploration-exploitation level.
\end{itemize}

\textbf{Trials Execution.} Figure~\ref{fig:trial-workflow} illustrates the workflow for executing trials. The trial executor runs a series of experiments until the desired objective is met. In each experiment, the optimizer (DE or BO) proposes new PID gains \((K_p, K_i, K_d)\). The trial executor applies these gains to the robot's PID controller and and command the robot to perform a 90-degree in-place rotation. After the rotation, the executor collects angular data from the robot's IMU sensor and sends it back to the optimizer, which checks whether the experiment meets the predefined constraints by calculating overshoot percentage and rise time. If both metrics lie within their respective boundaries, the experiment is marked as “accepted,” and the settling time is computed. The settling time serves as the primary optimization objective, which the framework aims to minimize. If the settling time becomes less than or equal to the specified threshold, the trial concludes and the logs are exported for further evaluation.

%% file: Sections/new_eval.tex
\begin{table*}[t]
\vspace{17px}

  \centering
  \caption{Experimental Results}
  \label{table:exp-results}
  \resizebox{\textwidth}{!}{%
    \begin{tabular}{@{}>{\centering\arraybackslash}m{3cm} >{\centering\arraybackslash}m{3cm} c c c c c@{}l}
      \toprule
      Robot Type & Exploration-Exploitation Configuration & Optimizer & Settling Time (ms) & Convergence Percentage & Rise Time (ms) & Overshoot Percentage  & Iteration\\
      \midrule
      \multicolumn{7}{c}{\textbf{Initial State 1}} & \\
      \cmidrule(lr){1-8}
      \multirow{2}{*}{Differential Drive} & \multirow{5}{*}{Balanced } & BO & 1118& 100\% & 421& 32.95 & 3\\[0.5em]
       &  & DE & 1406& 100\% & 631& 27.83 & 16\\[0.08cm] \cline{1-1}
      \multirow{2}{*}{Omnidirectional} &  & BO & 1225& 90\% & 553& 32.48 & 13\\[0.5em]
       &  & DE & 1535& 100\% & 658& 16.46 & 16\\[0.08cm] \midrule
      \multirow{2}{*}{Differential Drive} & \multirow{5}{*}{Exploration-Focused} & BO & 1378& 100\% & 432& 30.12 & 11\\[0.5em]
       &  & DE & 1418& 100\% & 452& 24.47 & 16\\[0.08cm] \cline{1-1}
      \multirow{2}{*}{Omnidirectional} &  & BO & 1514& 100\% & 553& 20.45 & 15\\[0.5em]
       &  & DE & 1552& 100\% & 553& 29.4 & 66\\[0.08cm] \midrule
      \multirow{2}{*}{Differential Drive} & \multirow{5}{*}{Exploitation-Focused } & BO & 1667& 100\% & 492& 32.88 & 19\\[0.5em]
       &  & DE & 1417& 100\% & 519& 31.11 & 31\\[0.08cm] \cline{1-1}
      \multirow{2}{*}{Omnidirectional} &  & BO & 1503& 80\% & 553& 20.45 & 15\\[0.5em]
       &  & DE & 1522& 100\% & 568& 29.09 & 61\\[0.08cm] \midrule
      \multicolumn{7}{c}{\textbf{Initial State 2}} & \\
      \cmidrule(lr){1-8}
      \multirow{2}{*}{Differential Drive} & \multirow{5}{*}{Balanced} & BO & 1151& 90\% & 565& 27.69 & 10\\[0.5em]
       &  & DE & 1451& 100\% & 485& 29.98 & 31\\[0.08cm] \cline{1-1}
      \multirow{2}{*}{Omnidirectional} &  & BO & 1246& 100\% & 590& 33.21 & 9\\[0.5em]
       &  & DE & 1479& 100\% & 538& 19.38 & 26\\[0.08cm] \midrule
      \multirow{2}{*}{Differential Drive} & \multirow{5}{*}{Exploration-Focused} & BO & 1421& 100\% & 517& 26.78 & 15\\[0.5em]
       &  & DE & 1488& 100\% & 487& 27.88 & 16\\[0.08cm] \cline{1-1}
      \multirow{2}{*}{Omnidirectional} &  & BO & 1420& 100\% & 553& 28.2 & 17\\[0.5em]
       &  & DE & 1501& 100\% & 553& 32.8 & 17\\[0.08cm] \midrule
      \multirow{2}{*}{Differential Drive} & \multirow{5}{*}{Exploitation-Focused} & BO & 1609& 90\% & 534& 33.06 & 17\\[0.5em]
       &  & DE & 1319& 100\% & 648& 29.31 & 11\\[0.08cm] \cline{1-1}
      \multirow{2}{*}{Omnidirectional} &  & BO & 1598& 80\% & 553& 28.2 & 17\\[0.5em]
       &  & DE & 1654& 100\% & 533& 28.41 & 56\\[0.08cm] \midrule

    \end{tabular}%
  }
\end{table*}

\begin{figure*}[t]
    \centering
    \includegraphics[width=0.8\textwidth]{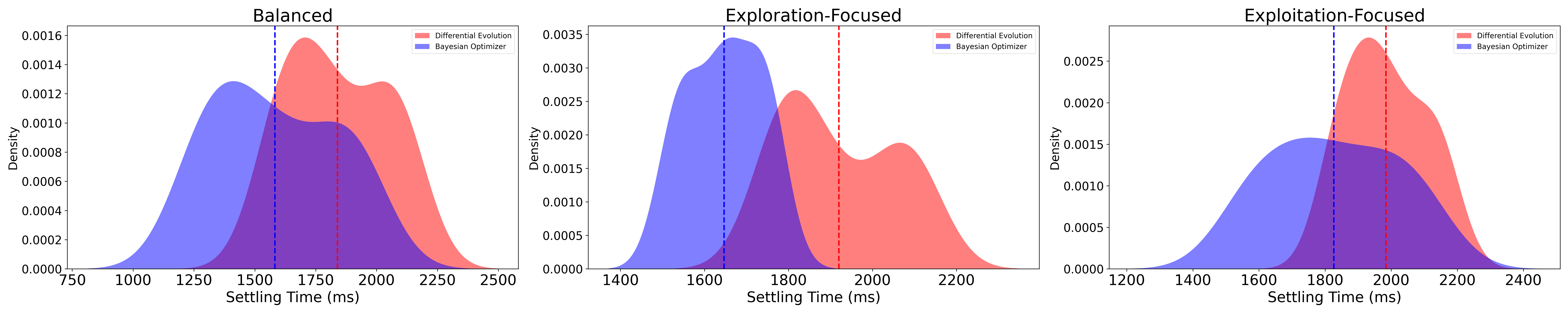}
    \caption{Settling-time distributions (KDE) for the omnidirectional robot under the three exploration--exploitation configurations. BO is shown in blue and DE in red; the dashed vertical lines indicate the mean settling times.}
    \label{fig:scout_kde}
\end{figure*}

\begin{figure*}[t]
    \centering
    \includegraphics[width=0.9\textwidth]{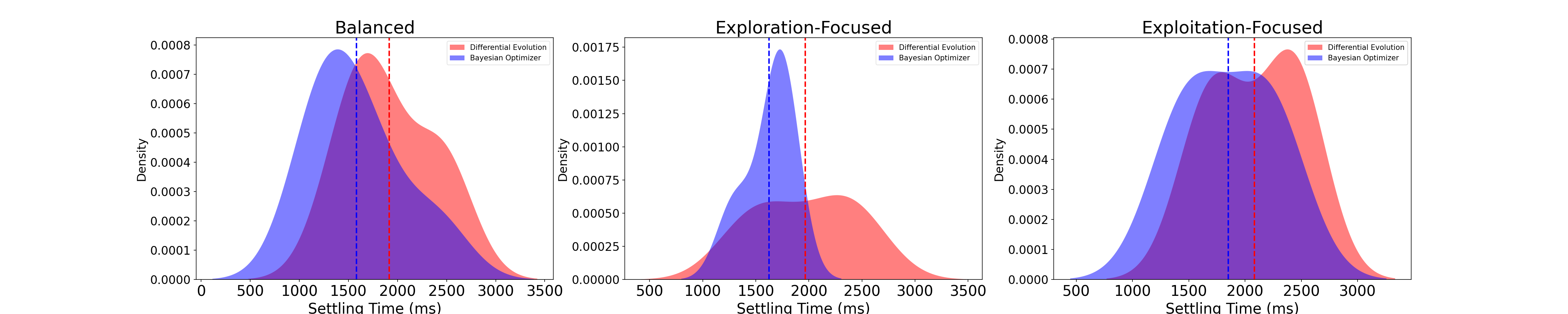}
    \caption{Settling-time distributions (KDE) for the differential drive robot under the three exploration--exploitation configurations. BO is shown in blue and DE in red; the dashed vertical lines indicate the mean settling times.}
    \label{fig:diffdrive_kde}
\end{figure*}

Twenty-four distinct trials were conducted, each repeated 10 times, resulting in a total of 240 runs. In each trial, the generated configuration was applied to both BO and DE and executed on both robot platforms. Table \ref{table:exp-results} reports the best performance achieved among the 10 runs in each trial, except for the “Convergence Percentage,” which indicates the proportion of runs that successfully converged. As explained in Section \ref{sec:methodology}, each trial represents experimenting using a unique configuration generated by the Configurations Generator, where each configuration represents a distinct pair of an init-state and exploration-exploitation level. Additionally, Figures \ref{fig:scout_kde} and \ref{fig:diffdrive_kde} illustrate the settling-time distributions for BO and DE across the different exploration–exploitation levels. The following subsections address the RQs introduced in Section \ref{sec:introduction} and discuss the key findings in detail.

\subsection{RQ1: Effect of Exploration–Exploitation on Convergence and Settling Time} When the differential drive robot used Initial State 1 under a Balanced exploration–exploitation setting, it consistently showed the fastest settling times and a perfect convergence rate. For instance, BO achieved 1118ms and DE reached 1406ms, illustrating that an even mix of exploration and exploitation can quickly guide the optimization to suitable parameter values. Shifting to either Exploration-Focused or Exploitation-Focused levels raised settling times slightly, yet DE tended to benefit more from heavier exploitation than BO did.

For the omnidirectional robot, Balanced exploration–exploitation again led to the best settling times under Initial State 1, although BO’s convergence rate dropped when the focus shifted to exploitation. This suggests that while an omnidirectional robot may profit from balanced tuning, heavy exploitation can sometimes overlook valid regions of the parameter space that BO struggles to recover.

\textbf{A key conclusion for RQ1} is that a Balanced exploration–exploitation strategy generally offers the most stable and rapid convergence. Although some scenarios demonstrate improved performance under either exploration or exploitation, balancing them tends to reduce the risk of poor convergence, especially for BO. In contrast, DE exhibits robust behavior across exploration-exploitation levels but can especially excel when set to high exploitation.

\subsection{RQ2: Influence of Initial State on Settling Time and Convergence} In the differential drive robot, Initial State 1 usually allowed faster settling times than Initial State 2, particularly under Balanced level. BO, for instance, improved from 1151ms with Initial State 2 to 1118ms with Initial State 1, indicating that a stronger proportional response from the outset can guide optimization more effectively. DE also showed reliable performance but benefited significantly if its initial conditions already aligned well with the task.

For the omnidirectional robot, changes in initial states did not dramatically alter its performance. While Initial State 1 combined with Balanced tuning gave BO an edge in settling time, that advantage came with reduced convergence reliability. This pattern indicates that the omnidirectional platform is inherently more tolerant of different initial PID values, although there is still some interaction between initial conditions and how exploration and exploitation are balanced.

\textbf{A key conclusion for RQ2} is that choosing the right initial state can substantially affect the PID auto-tuning results. When the initial gains already steer the system in a promising direction, the optimizer refines parameters more quickly. This effect becomes even more clear under exploitation-focused levels, where being closer to the global optimum accelerates convergence. In contrast to the differential drive robot, the omnidirectional's performance varies less across different initial states, suggesting that its geometry and motion characteristics make it less sensitive to small variations in the initial PID configuration.

\subsection{RQ3: Comparison of BO and DE for PID Auto-Tuning} BO commonly leads to achieving lower settling times in fewer iterations when exploration and exploitation are well balanced. For example, it reached 1118ms for the differential drive robot with Initial State 1, compared to 1406ms under DE, which highlights BO’s efficiency in honing in on high-potential parameter regions. However, when exploitation dominates, DE can surpass BO, as evidenced by DE’s 1319ms against BO’s 1609ms under Initial State 2. This outcome shows that a heavier focus on refining certain promising areas may benefit DE more, provided it can maintain feasible solutions along the way.

Another critical difference between the two methods is convergence reliability. DE maintained a perfect 100\% convergence rate across all trials, while BO occasionally failed to converge, especially in scenarios that reduced its capacity to explore broader parameter spaces. DE’s deterministic strategy ensures that feasible solutions eventually emerge, though it may need more iterations to reach the same level of fine-tuned performance that BO can achieve under optimal conditions.

\textbf{A key conclusion for RQ3} is that BO’s principal advantage is rapid improvement in settling time when exploration and exploitation are balanced effectively, but it is more sensitive to both parameter space complexity and initial conditions. DE is robust and achieves consistent convergence, showing particular strength under exploitation-heavy strategies, though it might require additional iterations to match or exceed BO’s best results. These findings suggest that the choice between BO and DE depends on the system’s need for guaranteed convergence (favoring DE), the importance of fast tuning (favoring BO in balanced scenarios), and the degree to which initial conditions are known to guide or hinder the optimization process.

\vspace*{0.55cm}

%% file: Sections/Related.tex
This section compares our proposed approach with the following threads of research: (a) Tuning PID controllers for mobile robotics, (b) DE for tuning PID controllers.and (c) BO for tuning PID controllers.

\textbf{Differential Evolution for tuning PID controllers.} DE is widely used to tune PID controllers in mobile robots due to its effectiveness in handling complex non-linear optimization problems. DE has been successfully applied across various robotic platforms, including differential drive, omnidirectional, and parallel robots, enhancing control performance and achieving faster convergence than traditional methods like the Teaching-Learning Based Optimization (TLBO) and Genetic Algorithms (GA) \cite{comparative_study_of_optimal_tuning_pid_controller_for_manipulator_robot_2023, yongju_pak__2022}. DE has outperformed conventional approaches, such as Ziegler-Nichols tuning, in optimizing PID parameters for trajectory tracking, disturbance attenuation, and precise control of robot kinematics \cite{peizhang_wu__2017, najah_yousfi_allagui__2021}. Its integration into advanced algorithms, such as fuzzy-PID control, further enhances trajectory accuracy and system responsiveness in challenging scenarios.

\textbf{Bayesian optimization for tuning PID controllers.}  BO has been applied especially those with nonlinear and underactuated systems like wheel mobile robots (WMR) \cite{nur_aisyah_syafinaz_suarin__2019}. The BO can be compared to other optimization techniques such as genetic algorithms (GA) and Bat Algorithm (BA). BO has been used to fine-tune PID parameters for AGVs and WMRs, respectively, demonstrating significant improvements in path-following accuracy and system performance \cite{nur_aisyah_syafinaz_suarin__2019}\cite{EvolutiveTuningOptimization}.

\textbf{Tuning PID controllers for mobile robotics}. Traditional manual tuning methods, such as Ziegler-Nichols, can be time-consuming, can't be performed in certain cases, or may not yield the best results for complex mobile robotic systems. Recent studies have explored various optimization techniques for PID controller tuning. Various studies have optimized PID parameters using different algorithms. For DC motor control, the Bees Algorithm, Particle Swarm Optimization (PSO), and Teaching-Learning-Based Optimization (TLBO) were used in MATLAB to improve performance metrics such as overshoot and settling time \cite{DCMotorPIDTuning}. Grey Wolf Optimizer (GWO) optimized PID tuning for a quadruped robot leg in Simulink, compared to the Genetic Algorithm (GA) and PSO \cite{GreyWolfOptimizerAlgorithm}. The Social Spider optimization improved speed control in wheeled robots under disturbances \cite{WheeledMobileRobotbySocialSpider}. GA was used for PID tuning in Automated Guided Vehicles (AGVs), enhancing the path-following accuracy \cite{EvolutiveTuningOptimization}. Deep reinforcement learning outperformed fuzzy logic-based PI controllers in robotic drivers \cite{DeepReinforcementLearning}. An improved PSO algorithm improved PID tuning stability and convergence in different drive robotics \cite{PIDParameterTuningBasedOnImprovedParticleSwarmOptimization}. 

%% file: Sections/Conclusion.tex
This study presented a novel framework for PID auto-tuning, utilizing BO and DE to streamline experimentation and analyze the impact of initial system states and the exploration-exploitation trade-off on convergence dynamics. Experimental results showed that a balanced exploration-exploitation approach led to the fastest convergence and lowest settling times, especially when paired with well-chosen initial PID states. DE consistently ensured robust convergence, making it a stable choice, while BO excelled in reducing settling time but was more sensitive to initial configurations and parameter complexity. The key findings and insights contribute to broader fields where fine-tuned control mechanisms are essential for improving robotic performance. Future work can explore hybrid approaches that integrate BO’s efficiency with DE’s robustness, potentially leading to even more effective PID tuning methodologies. Additionally, extending this framework to multi-objective optimization scenarios and real-time adaptation could further enhance its applicability across diverse CPS platforms. 